\documentclass{article}

\usepackage[numbers]{natbib}
\usepackage[final]{neurips_2021}
\usepackage[utf8]{inputenc} 
\usepackage[T1]{fontenc}    
\usepackage{hyperref}       
\usepackage{url}            
\usepackage{booktabs}       
\usepackage{amsfonts}       
\usepackage{nicefrac}       
\usepackage{microtype}      
\usepackage{xcolor}         
\usepackage{amsmath}
\usepackage{subfig}
\usepackage{graphicx}
\usepackage{titlesec}
\titlespacing*{\section}{0pt}{0.25\baselineskip}{0.25\baselineskip}
\titlespacing*{\subsection}{0pt}{0.2\baselineskip}{0.2\baselineskip}

\title{Real-Time Monitoring of User Stress, Heart Rate, \\ and Heart Rate Variability on Mobile Devices}

\author{%
  Peyman Bateni$^1$ and Leonid Sigal$^{1,2,3,4}$ \\
  $^1$Beam AI, $^2$University of British Columbia, $^3$Vector Institute, $^4$CIFAR AI Chair  \\
  Correspondence to \texttt{pbateni@beamhealth.ai} \\
}

\begin{document}

\maketitle

\vspace{-0.15in}
\begin{abstract}
Stress is considered to be the epidemic of the 21st-century \cite{Fink2016_StressEpidemic}. Yet, mobile apps cannot directly evaluate the impact of their content and services on user stress. We introduce the Beam AI SDK to address this issue. Using our SDK, apps can monitor user stress through the selfie camera in real-time. Our technology extracts the user’s pulse wave by analyzing subtle color variations across the skin regions of the user’s face. The user’s pulse wave is then used to determine stress (according to the Baevsky Stress Index), heart rate, and heart rate variability. We evaluate our technology on the UBFC dataset, the MMSE-HR dataset, and Beam AI's internal data. Our technology achieves 99.2\%, 97.8\% and 98.5\% accuracy for heart rate estimation on each benchmark respectively, a nearly twice lower error rate than competing methods. We further demonstrate an average Pearson correlation of 0.801 in determining stress and heart rate variability, thus producing commercially useful readings to derive content decisions in apps. Our SDK is available for use\footnote{Visit \url{beamhealth.ai} for more information. The Beam AI SDK (iOS) files can be accessed at \url{github.com/beamai/BeamAISDK-iOS}. SDK access keys can be obtained by signing up at \url{dashboard.beamhealth.ai}.}.
\end{abstract}

\begin{figure}[!b]
    \centering
    \includegraphics[width=\textwidth]{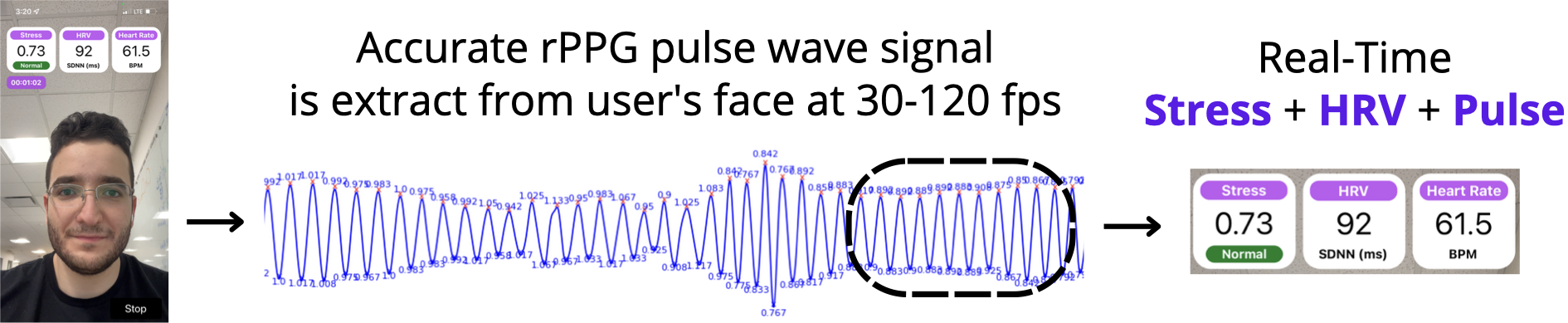}
    \caption{{\bf Overview of the Beam AI SDK.} First, the user’s pulse wave is extracted by processing subtle color variations across the skin regions of the user’s face. The user’s pulse is then processed by the SDK's proprietary peak detection algorithm which produces the inter-beat intervals used to determine the user's stress (according to Baevsky Stress Index), heart rate and heart rate variability.}
    \label{fig:sdk-overview}
\end{figure}

\section{Introduction}

It is estimated that over \$1 trillion in economic activity is lost due to stress every year \cite{WHO_GDPLossStress}. Globally, 275 million people suffer from stress disorders \cite{Fleming2019_StressDisorders}, and the COVID-19 pandemic has significantly exacerbated the severity and prevalence of stress-induced illnesses \cite{Cost22_CovidCanadaMentalHealth, GarcíaFernández22_SpainCovidHealthWorkers, RavensSieberer22_CovidGermany}. Today, over 6.5 billion smartphone devices are in use \cite{Turner22_Smartphones}, and the average smartphone user spends 3 hours and 15 minutes on their device every day \cite{Howarth22_SmartphoneUsage}. This has put smartphone apps at the center of the global stress crisis. Social media apps have been associated with significant increases in stress and the overall worsening of mental health in users \cite{McHugh18_SocialMediaTeens,Ngien22_SocialMediaPandemic}. Mobile games can cause severe stress and depression  when played excessively \cite{Wang19_GameMentalHealth}. To manage, reduce and prevent stress in smartphone apps, it is essential to enable apps to monitor user stress accurately and in real-time.

Stress monitoring is especially important for mental health apps. The COVID-19 pandemic aggravated the need for mental health services, significantly worsening the existing shortage of mental health providers \cite{Caron21_ProvidersStruggle}. Today, an estimated 20 thousand mental health apps have been released to address the ever-growing need for mental health solutions \cite{Auxier21_MentalHealthApps}. These apps provide a wide range of content and services to help with stress, anxiety, mood, and the overall mental health of their users. 
In order to validate their content and services, mental health apps need to understand the impact their content and services have on the user stress. However, existing procedures primarily rely on user studies that are slow, expensive, and limited in scope and accuracy.

A growing number of mental health apps have begun experimenting with using the user's heart rate variability, as measured over time through smart wearables with bio-metric capabilities, to evaluate the impact of their content and services. Heart rate variability is broadly considered to be the most accurate quantitative physiological measure of stress \cite{BaevskyStressIndex, Kim18_StressHRV}. Despite positive results, there are three major problems with the use of wearables. First, 70-90\% of users do not have wearables. Second, even users that have wearables, don’t always wear them or have them readily available. Third, many wearables (including the Apple Watch) do not allow third-party apps to continuously monitor the user’s heart rate variability, despite having the hardware capabilities.

We introduce the Beam AI SDK to enable apps to monitor user stress through the selfie camera in real-time. The SDK first extracts the user’s pulse wave by analyzing subtle color variations across the skin regions of the user’s face. The user’s pulse is then used to determine stress, heart rate, and heart rate variability. To calculate stress, we analyze the variability of the user's heartbeats according to the Baevsky Stress Index \cite{BaevskyStressIndex, Kim18_StressHRV}. Our SDK is readily available (requiring no additional hardware), operates fully on-device (with the user data never leaving the phone), is computationally efficient, and can run simultaneously in the background during (any) app usage.

Our contributions are as follows:
\begin{itemize}
    \item We present the Beam AI SDK which enables apps to monitor user stress in real-time, and we provide two demo apps (Beam AI Lite and Beam AI Browser) built with the Beam AI SDK.
    \item We empirically evaluate the efficacy of the Beam AI SDK on UBFC \cite{Bobbia17_UBFC}, MMSE-HR \cite{Zhang16_MMSE} and Beam AI's internal datasets. We show that our core technology is able to achieve nearly twice better accuracy when estimating user heart rate. We further demonstrate an average Pearson correlation of 0.801 in determining stress and heart rate variability as compared to gold-standard readings.
\end{itemize}

The remainder of this paper is structured as follows. In Section \ref{sec:technology}, we outline our core technology and provide high-level technical details. We furthermore provide an overview of our demo apps. In Section \ref{sec:empirical-evaluation}, we provide an extensive empirical evaluation of our core technology. In Section \ref{sec:discussion}, we summarize our work and provide plans for future development.

\begin{table}[t]
    \centering
    \tabcolsep=0.67cm
    \begin{tabular}{lcccc}
        {} & \multicolumn{4}{c}{\textbf{UBFC Benchmark (30 fps)}} \\
        Model & MAE$\downarrow$ & MAPE$\downarrow$ & RMSE$\downarrow$ & $\rho$ $\uparrow$ \\
        \midrule
        Beam AI SDK & \textbf{0.65} & \textbf{0.77\%} & 1.98 & \textbf{0.99} \\
        \midrule
        EfficientPhys-T1 \cite{EfficientPhys_Liu2021} & 2.08 & 2.53\% & 4.91 & 0.96 \\
        TS-CAN \cite{TSCAN_Liu2020} & 1.70 & 1.99\% & 2.72 & \textbf{0.99} \\
        EfficientPhys-C \cite{EfficientPhys_Liu2021} & 1.14 & 1.16\% & \textbf{1.81} & \textbf{0.99} \\
        POS \cite{POS_Wang2016} & 3.52 & 3.36\% & 8.38 & 0.90 \\
        CHROM \cite{CHROM_DeHaan2013} & 3.10 & 3.83\% & 6.84 & 0.93 \\
        ICA \cite{ICA_Poh2011} & 4.39 & 4.30\% & 11.60 & 0.82 \\
    \end{tabular}
    \vspace{0.05in}
    \caption{Heart rate estimation on UBFC \cite{Bobbia17_UBFC} according to the widely evaluated experimental setting of Liu et al. \cite{TSCAN_Liu2020, MetaPhys_Liu2021, EfficientPhys_Liu2021}. Values in bold indicate state of the art performance. MAE for heart rate estimation is measured in beats per minute.}
    \label{tab:liu-eval-ubfc}
    \vspace{-0.2in}
\end{table}

\section{Technology}
\label{sec:technology}

An overview of the core technology inside the Beam AI SDK is shown in Figure \ref{fig:sdk-overview}. The SDK consists of three modules: the pulse extractor, the inter-beat interval processor, and the biometric estimator.

\begin{itemize}
    \item \textbf{Pulse Extractor:} A camera session is managed internally within the SDK. When the user's face is present, the user’s pulse wave is continuously extracted by processing subtle color variations across the skin regions of the face. This is completed using Beam AI's proprietary real-time remote photoplethysmography technology and is updated with every new frame.
    \item \textbf{Inter-Beat Interval Processor:} As the user's pulse wave is updated, it is reprocessed to identify any new pulse peaks. If a new pulse peak is detected, then it is used to determine the inter-beat interval that is between this peak and the previous pulse peak. This inter-beat interval is calculated and added to the user's set of sequentially detected inter-beat intervals.
    \item \textbf{Biometric Estimator:} To produce the reading, the inter-beat intervals constituting the last $t$ seconds of the user's pulse are used to determine heart rate, heart rate variability, and stress of the user. Note that the window over which these readings are calculated ($t$) and the frequency at which the readings are re-estimated are hyperparameters that are defined when the Beam aI SDK is initialized.
    \begin{itemize}
    \item \textbf{Heart Rate:} Heart rate describes the number of beats per minute observed over a window of time. For a window of time, the inter-beat intervals are extracted from the pulse wave. Given a set of inter-beat intervals $\{\text{IBI}_i\}$, pulse wave is calculated in beats-per-minute by:
    \begin{equation}
        \text{Pulse}(\{\text{IBI}_i\}) = \frac{60}{\frac{1}{N} \sum_{i=1}^N \text{IBI}_i}
    \end{equation}
    \item \textbf{Heart Rate Variability:} Heart rate variability is concerned with analyzing not the average beat interval length but instead how variable the beat intervals are over a span of time. We report heart rate variability according to the standard deviation of the IBI of normal sinus beats (SDNN) which is also used by the Apple Watch. We measure SDNN in milliseconds.
    \item \textbf{Stress:} We determine stress according to the Baevsky Stress Index. Baevsky is a complex heart rate variability metric shown to correlate best with physiological stress \cite{BaevskyStressIndex}. Given a set of inter-beat intervals $\{\text{IBI}_i\}$, we calculate Baevsky stress according to the formula below.
    \vspace{0.1in}
    \begin{equation}
        SI(\{\text{IBI}_i\}) = \frac{\text{amp}(\text{mod}_\text{50ms}(\{\text{IBI}_i\}))}{ 2 * \text{mod}_\text{50ms}(\{\text{IBI}_i\}) * 3.92 * \text{SDNN}(\{\text{IBI}_i\})}
        \vspace{0.1in}
    \end{equation}
    where the $\text{mod}$ function takes the mode of the histogram of the inter-beat intervals binned in 50 ms long bins. The $\text{amp}$ function returns the amplitude of the mode of the histogram as defined by the percentage of inter-beat intervals in that specific bin. Note that conventionally, instead of the SDNN term, the difference between the longest and shortest intervals is used, corresponding to the full range of inter-beat intervals observed. However, in our work we use the $3.92 * \text{SDNN}(\{\text{IBI}_i\})$. This corresponds to the range spanning 95\% of interval samples within the set (i.e. 1.96 standard deviations in either direction of the mean). This was experimentally observed to provide readings more robust to noise that arises from the misclassification of a single inter-beat interval. 
\end{itemize}
\end{itemize}

The Beam AI SDK can estimate user heart rate anywhere between 39 and 210 beats per minute. This provides an extensive coverage of heart rhythms generally observed in humans. By comparison, the optical sensor on the Apple Watch supports a range of 30 to 210 beats per minute \cite{AppleSupport2022}. The heart-rate sensor on FitBit devices detect a range of 30 to 220 beats per minute \cite{FitBitHealthSolutions2022}. 

\subsection{Privacy and On-Device Processing} 

All processing within the Beam AI SDK happens on the mobile device. This is to ensure maximal user privacy as video data, especially with the user's face present, contains personal information. In the Beam AI SDK, video, face, and biometric data never leave the device. This includes any intermediary artifacts that arise from processing. Furthermore, we do not save video or face data on the device, but instead, only maintain the most recent frame in memory when processing new frames. This process takes 0.5 milliseconds (see Section \ref{sec:processing-speed}) and the frame is immediately deleted after processing. Biometric data is also not maintained on device for an extended period of time. 

All data, including intermediary artifacts, are erased once monitoring by the SDK stops. The Beam AI SDK does not collect or save any data externally or on device. That being said, third-party developers using the SDK may access the video data when using the SDK to provide a preview and can access biometric readings of the user during monitoring, and are therefore responsible for handling said data in accordance with desired preservation of user privacy. 

\subsection{Demo Apps} 

In addition to directly using the Beam AI SDK, you can test our technology using our demo apps that have been built using the Beam AI SDK.

\begin{table}[t]
    \centering
    \tabcolsep=0.67cm
    \begin{tabular}{lcccc}
        {} & \multicolumn{4}{c}{\textbf{MMSE-HR Benchmark (25 fps)}} \\
        Model & MAE$\downarrow$ & MAPE$\downarrow$ & RMSE$\downarrow$ & $\rho$ $\uparrow$ \\
        \midrule
        Beam AI SDK & \textbf{1.72} & \textbf{2.24\%} & \textbf{4.03} & \textbf{0.95} \\
        \midrule
        EfficientPhys-T1 \cite{EfficientPhys_Liu2021} & 3.04 & 3.91\% & 5.91 & 0.92 \\
        TS-CAN \cite{TSCAN_Liu2020} & 3.04 & 3.41\% & 6.55 & 0.89 \\
        EfficientPhys-C \cite{EfficientPhys_Liu2021} & 3.48 & 4.02\% & 7.21 & 0.86 \\
        POS \cite{POS_Wang2016} & 3.79 & 4.28\% & 8.47 & 0.82 \\
        CHROM \cite{CHROM_DeHaan2013} & 3.61 & 4.50\% & 7.43 & 0.85 \\
        ICA \cite{ICA_Poh2011} & 7.96 & 9.20\% & 14.02 & 0.51 \\
    \end{tabular}
    \vspace{0.05in}
    \caption{Heart rate estimation on MMSE-HR \cite{Zhang16_MMSE} according to the widely evaluated experimental setting of Liu et al. \cite{TSCAN_Liu2020, MetaPhys_Liu2021, EfficientPhys_Liu2021}. Values in bold indicate state of the art performance. MAE for heart rate estimation is measured in beats per minute.}
    \label{tab:liu-eval-mmse}
    \vspace{-0.2in}
\end{table}

\begin{figure}
    \centering
    \subfloat[]{{\includegraphics[width=0.32\textwidth]{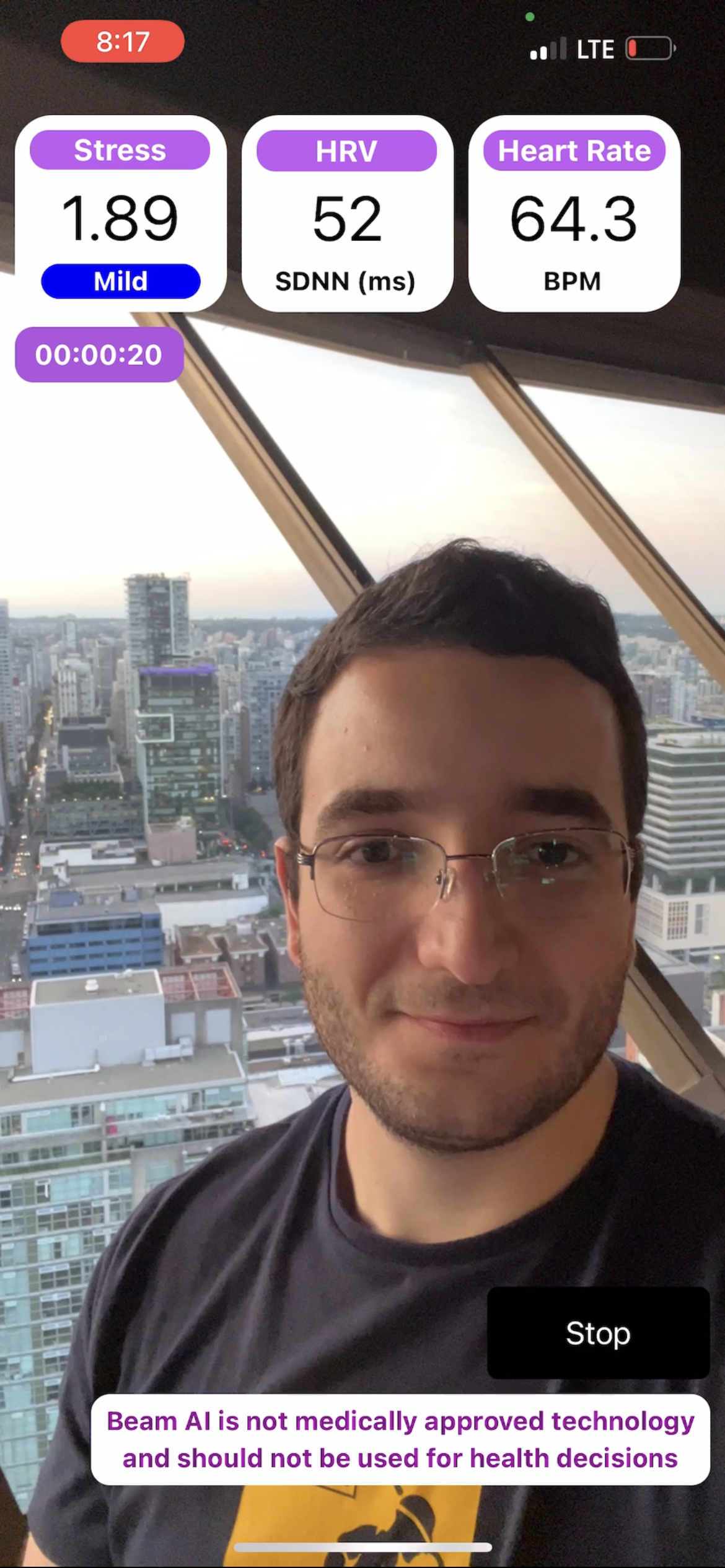} }}
    \subfloat[]{{\includegraphics[width=0.32\textwidth]{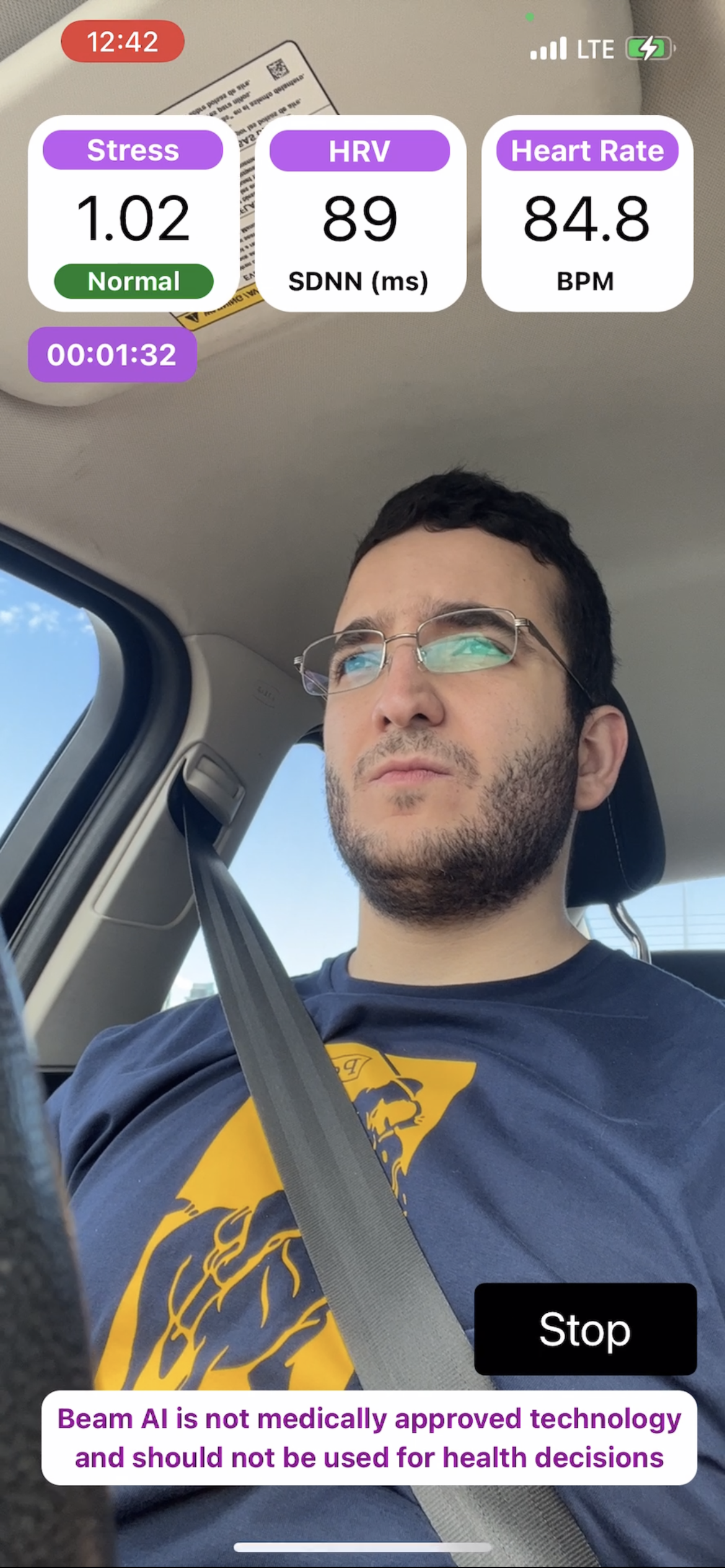} }}
    \subfloat[]{{\includegraphics[width=0.32\textwidth]{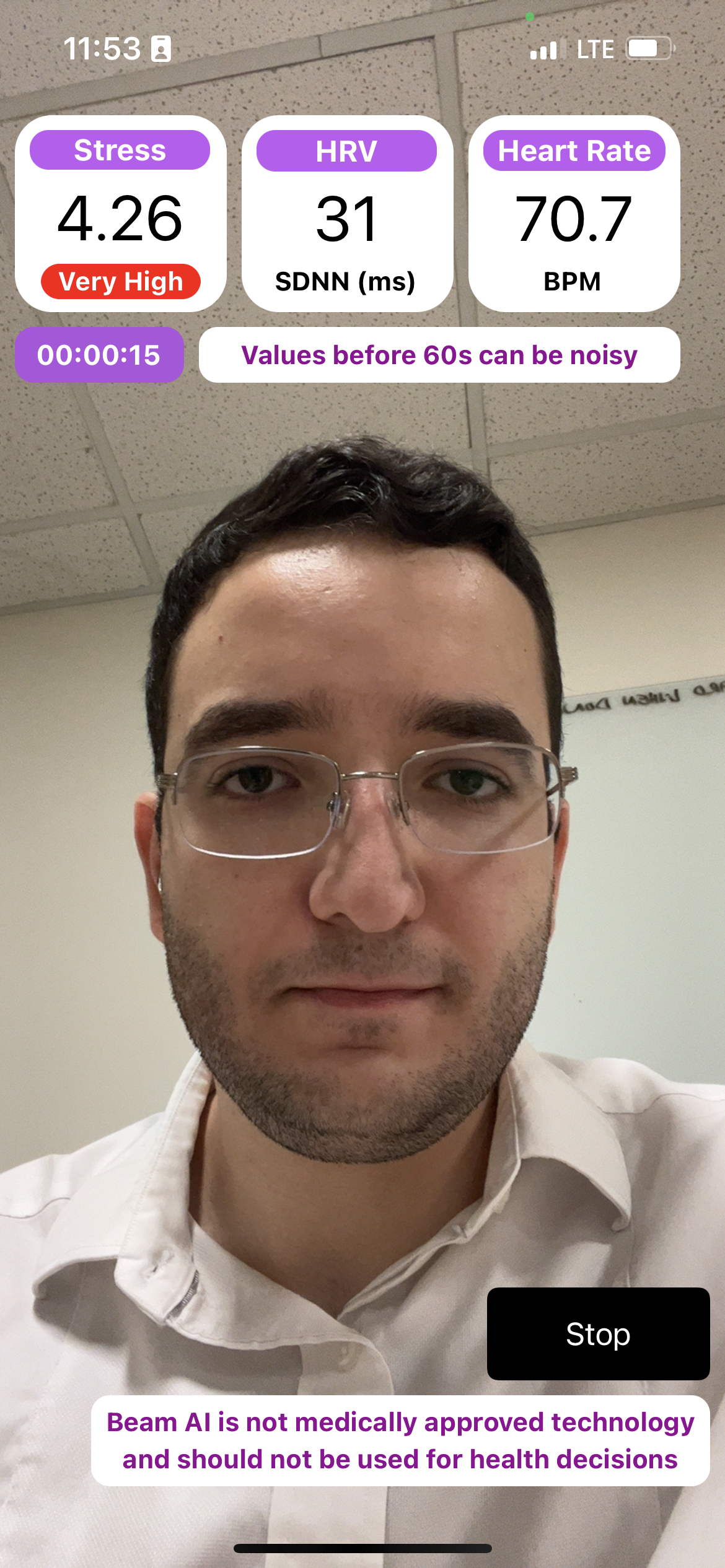} }} \\
    \subfloat[]{{\includegraphics[width=0.32\textwidth]{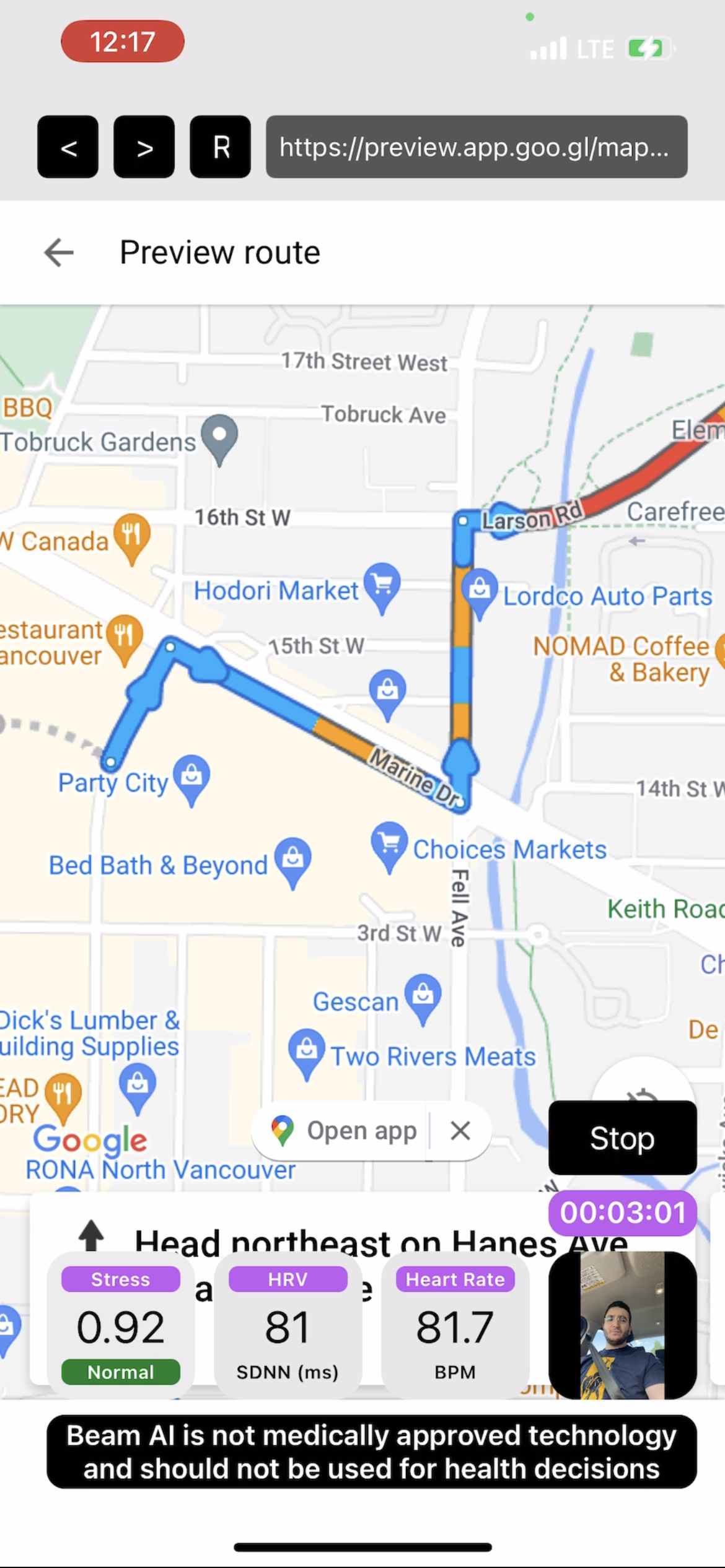} }}
    \subfloat[]{{\includegraphics[width=0.32\textwidth]{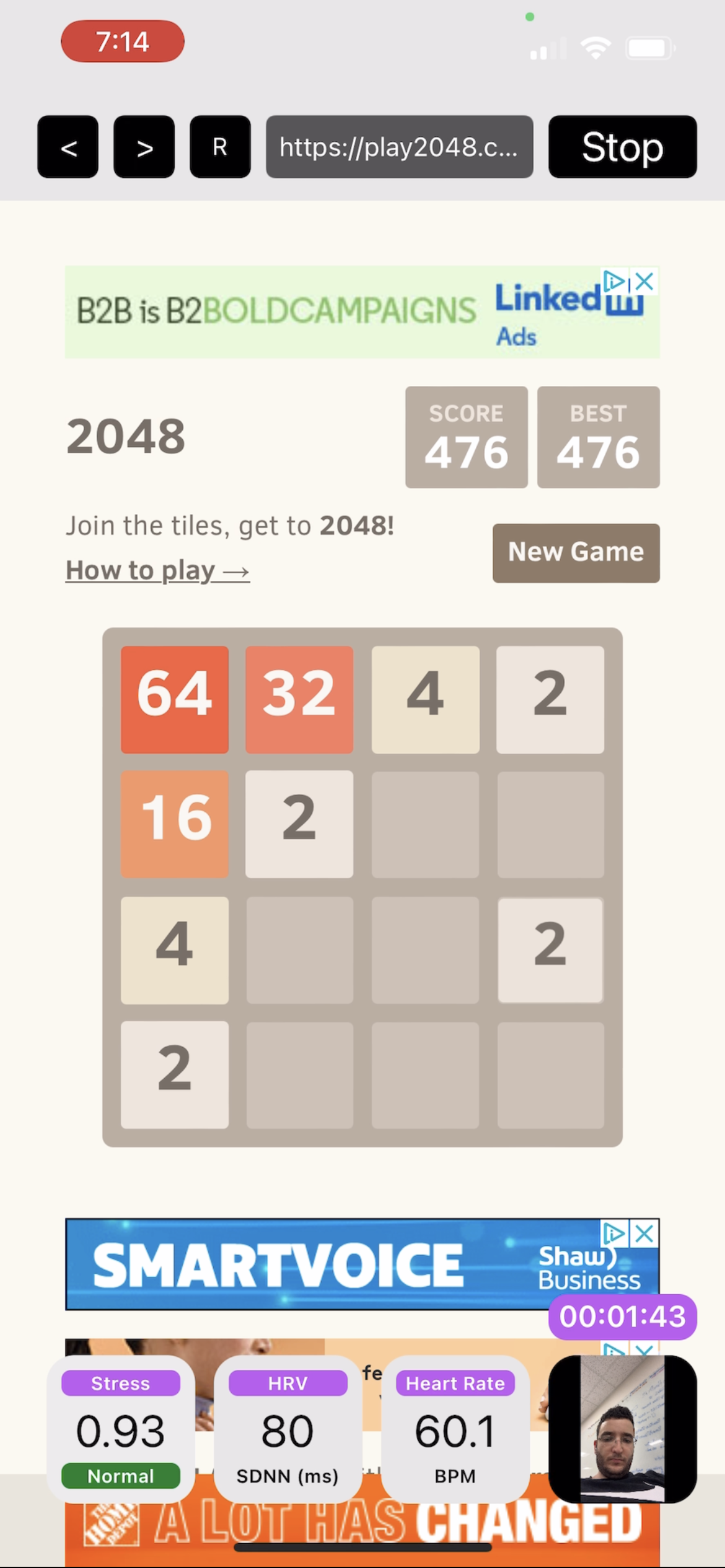} }}
    \subfloat[]{{\includegraphics[width=0.32\textwidth]{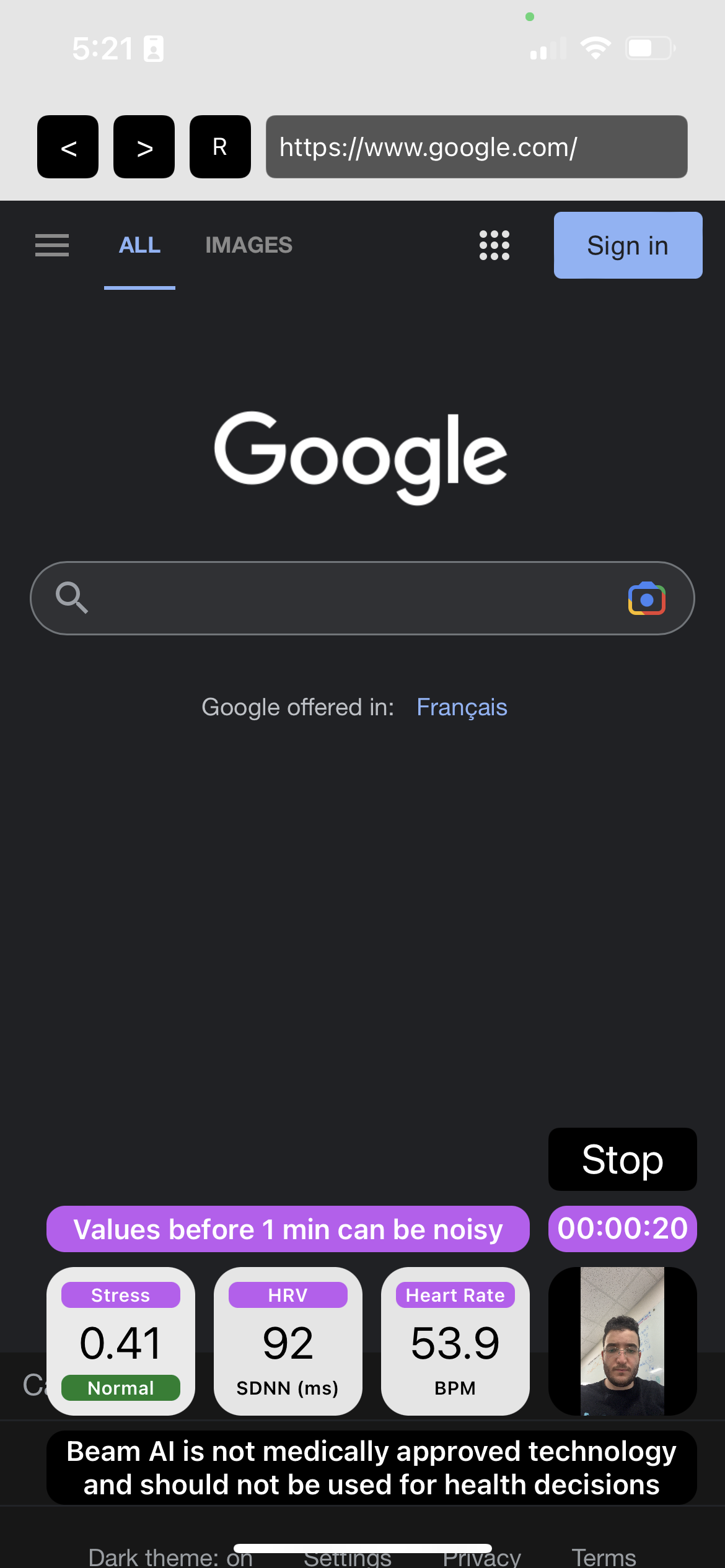} }}
    \caption{Screenshots of Beam AI Lite (a-c) and Beam AI Browser (d-f) apps built with the Beam AI SDK. Beam AI Lite demonstrates the core stress, heart rate, and heart rate variability monitoring technology provided by the Beam AI SDK. Beam AI Browser demonstrates passive stress monitoring with the Beam AI SDK as the user browses the web.}
    \label{fig:screenshots}
\end{figure}

\begin{itemize}
    \item \textbf{Beam AI Lite} demonstrates the core stress, heart rate, and heart rate variability monitoring technology in the Beam AI SDK. See Figure \ref{fig:screenshots}-a, \ref{fig:screenshots}-b and \ref{fig:screenshots}-c for screenshots of Beam AI Lite. Beam AI Lite is available on App Store\footnote{Download Beam AI Lite from \url{apps.apple.com/ca/app/beam-ai-lite/id1629758948}.} and its source code is available on GitHub\footnote{See Beam AI Lite's source code at \url{github.com/beamai/BeamAILite-iOS}.}.
    \item \textbf{Beam AI Browser} demonstrates passive stress monitoring with the Beam AI SDK as the user browses the web. See Figure \ref{fig:screenshots}-d, \ref{fig:screenshots}-e and \ref{fig:screenshots}-f for screenshots of Beam AI Browser. Beam AI Browser is available on App Store\footnote{Download Beam AI Browser from \url{apps.apple.com/ca/app/beam-ai-browser/id1629793784}.} and its source code is available on GitHub\footnote{See Beam AI Browser's source code at \url{github.com/beamai/BeamAIBrowser-iOS}.}.
\end{itemize}

\section{Empirical Evaluation}
\label{sec:empirical-evaluation}

Our empirical evaluation is structured as follows. In Section \ref{sec:definitions}, we define the evaluation metrics. In Section \ref{sec:prior-lit}, we evaluate our core technology on the widely reported experimental setting of Liu et al.\cite{TSCAN_Liu2020, EfficientPhys_Liu2021, MetaPhys_Liu2021} and discuss heart rate estimation results on UBFC \cite{Bobbia17_UBFC} and MMSE-HR \cite{Zhang16_MMSE} benchmarks. In Section \ref{sec:ubfc-clean}, we evaluate stress, heart rate and heart rate variability estimation on the UBFC \cite{Bobbia17_UBFC} benchmark using hand-verified accurate pulse peaks for ground truth. In Section \ref{sec:iphone-data}, we evaluate continuous monitoring of stress, heart rate and heart rate variability using Beam AI's internal data that consists of a 20-minute passive recording of Peyman Bateni on an iPhone 13 device. We compare these results to synchronized estimates from an Apple Watch and gold-standard readings from a Polar H10 chest strap.

\subsection{Metric Definitions}
\label{sec:definitions}

We use the following metrics for evaluating the performance of our technology.

\begin{itemize}
    \item \textbf{Mean Average Error (MAE):} For a set of predicted values $\{\hat{y}_i\}$ and corresponding ground truth target values $\{y_i\}$, MAE is defined to be the mean average error between the predicted and target values. It is calculated according to
    \begin{equation}
        \text{MAE}(\{\hat{y}_i\}, \{y_i\}) = \frac{1}{N} \sum_i |\hat{y}_i - y_i|\textit{.}
    \end{equation}
    \item \textbf{Mean Average Percentage Error (MAPE):} MAPE is an extension of MAE that measures average error as a percentage of the target ground truth value, providing a reasonable estimate of percentage error. It is calculated according to
    \begin{equation}
        \text{MAE}(\{\hat{y}_i\}, \{y_i\}) = \frac{1}{N} \sum_i \frac{|\hat{y}_i - y_i|}{y_i} \times 100\% \textit{.}
    \end{equation}
    \item \textbf{Root Mean Squared Error (RMSE):} RMSE is a measure of average square error. In RMSE, large differences between target and predicted values are amplified by the square operation. As a result, it focuses more on cases where there are major differences between predicted and target values. It is calculated according to
    \begin{equation}
        \text{MAE}(\{\hat{y}_i\}, \{y_i\}) = \frac{1}{N} \sum_i (\hat{y}_i - y_i)^2 \textit{.}
    \end{equation}
    \item \textbf{Pearson Correlation:} Lastly, we use the Pearson correlation between the predicted values $\{\hat{y_i}\}$ and corresponding ground truth target values $\{y_i\}$ to measure how well-correlated our estimated stress, heart rate, and heart rate variability readings are.
\end{itemize}

\begin{table}[t]
    \centering
    \tabcolsep=0.5cm
    \begin{tabular}{lrrrr}
        {} & \multicolumn{4}{c}{\textbf{UBFC Benchmark (30 fps)}} \\
        {} & \multicolumn{4}{c}{\textbf{with Manually Hand-Verified Peaks}} \\
        Reading & MAE$\downarrow$ & MAPE$\downarrow$ & RMSE$\downarrow$ & $\rho$ $\uparrow$ \\
        \midrule
        Heart Rate {\small (Beats Per Minute)} & 0.318 & 0.32\% & 0.424 & 0.999  \\
        Heart Rate Variability {\small (SDNN ms)} & 11.125 & 20.26\% & 14.412 & 0.841  \\ 
        Stress {\small (Baevsky Stress Index)} & 0.973 & 44.43\% & 1.294 & 0.730 \\
    \end{tabular}
    \vspace{0.05in}
    \caption{Comparing stress, heart rate, and heart rate variability estimation with the Beam AI SDK on videos from the UBFC \cite{Bobbia17_UBFC} benchmark to the ground truth estimates from a gold-standard pulse sensor with manually hand-verified pulse peaks.}
    \vspace{-0.2in}
    \label{tab:ubfc-eval-with-manual-data}
\end{table}

\subsection{Evaluation on Standard Public Benchmarks}
\label{sec:prior-lit}

We first evaluate our technology in the experimental setting of Liu et al. \cite{TSCAN_Liu2020} for heart rate estimation on the UBFC \cite{Bobbia17_UBFC} and MMSE-HR \cite{Zhang16_MMSE} benchmarks which are widely used in the academic literature \cite{TSCAN_Liu2020, EfficientPhys_Liu2021, MetaPhys_Liu2021}.

\subsubsubsection{\textbf{3.2.1 \hspace{0.075in} Evaluation on the UBFC \cite{Bobbia17_UBFC} Benchmark}}

\label{sec:ubfc}
\textbf{Benchmark:} UBFC \cite{Bobbia17_UBFC} is a dataset of 42 uncompressed 8-bit RGB videos from 42 subjects with a resolution of 640x480 recorded at 30fps. Each recording is accompanied by a synchronized pulse  wave signal with a sample rate of 30Hz. During the recording, the subjects are sat down at a 1-meter distance from the camera and asked to solve a puzzle on a computer device located below the camera. 

\textbf{Ground Truth Signal:} Following \cite{EfficientPhys_Liu2021}, a 2nd-order Butterworth \cite{Selesnick96_Butterworth} filter that excludes frequencies outside of 0.75Hz (corresponding to 45 beats per minute) and 2.5Hz (corresponding to 150 beats per minute) is applied to the synchronized pulse wave signal of each video. Then, pulse peaks are extracted from the resulting signal using the standard peak detection function of SciPy \cite{2020SciPy-NMeth} to produce inter-beat intervals that are used to estimate the ground truth heart rate value for each video.

\textbf{Predictions by the Beam AI SDK:} For each video, the Python implementation of the ``Pulse Processor'' from our SDK produces a high-quality pulse wave signal from the subject's face. A Python implementation of the ``Inter-Beat Interval Processor'' then generates the inter-beat intervals from this pulse wave signal with additional corrective post-processing steps according to our peak detection algorithm. The resulting inter-beat intervals for the whole video are then used to estimate the heart rate for the video. Note that our technology has built-in bandpass filtering capabilities that accommodate a range of heart rates between 39 beats per minute and 210 beats per minute which is wider than the 2nd-order Butterworth \cite{Selesnick96_Butterworth} filter applied to the ground truth readings.

\textbf{Results:} The results are reported in Table \ref{tab:liu-eval-ubfc} and compared to competing methods. As shown, we achieve an MAE of 0.65 beats per minute and an MAPE of 0.77\%, demonstrating state of the art accuracy well within a single beat per minute from the ground truth readings. Furthermore, we achieve a near-perfect Pearson correlation score, indicating that increases and decreases in heart rate readings from our technology directly follow the trends observed using the gold-standard ground truth heart rate measurements.

\begin{figure}[t]
    \centering
    \subfloat[Sample 1 - Pulse Wave Extracted by the Beam AI SDK vs. Gold-Standard]{{\includegraphics[width=0.83\textwidth]{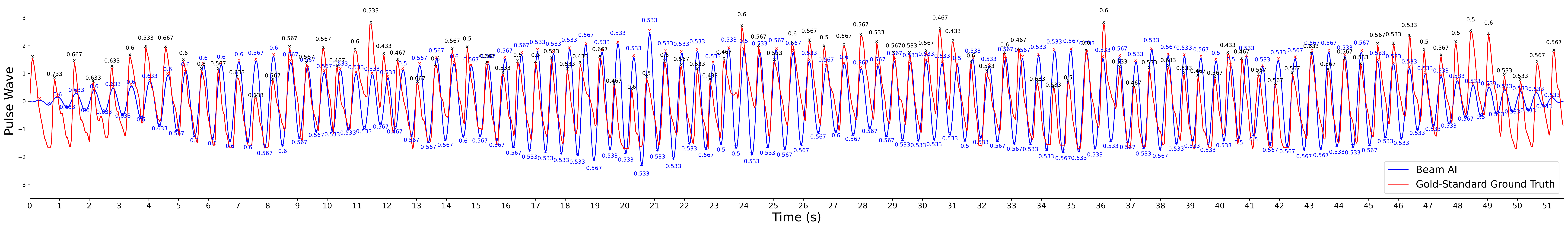} }} \\
    \vspace{-0.1in}
    \subfloat[Sample 2 - Pulse Wave Extracted by the Beam AI SDK vs. Gold-Standard]{{\includegraphics[width=0.83\textwidth]{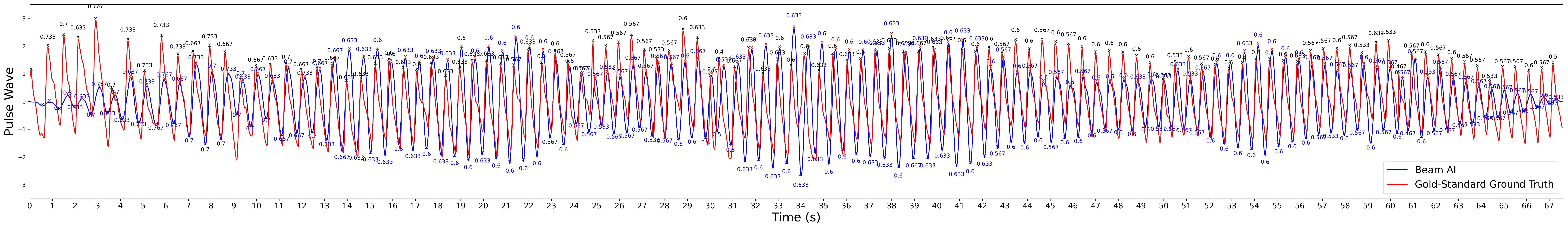} }} \\
    \vspace{-0.1in}
    \subfloat[Sample 3 - Pulse Wave Extracted by the Beam AI SDK vs. Gold-Standard]{{\includegraphics[width=0.83\textwidth]{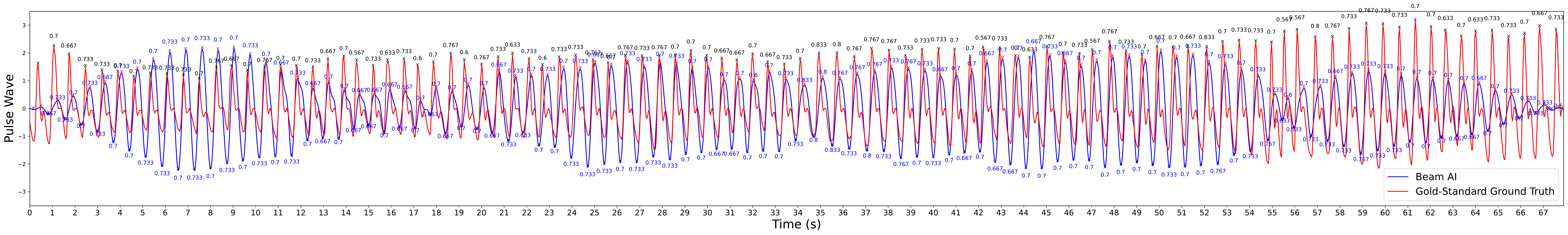} }} \\
    \vspace{-0.1in}
    \subfloat[Sample 4 - Pulse Wave Extracted by the Beam AI SDK vs. Gold-Standard]{{\includegraphics[width=0.83\textwidth]{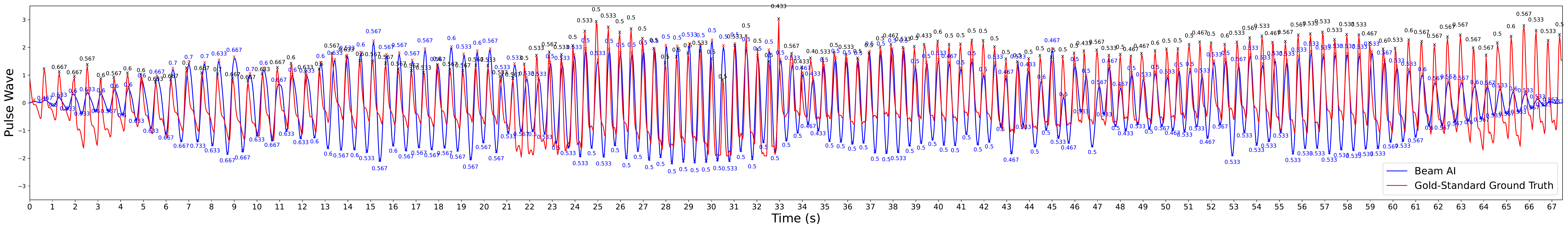} }} \\
    \vspace{-0.1in}
    \subfloat[Sample 5 - Pulse Wave Extracted by the Beam AI SDK vs. Gold-Standard]{{\includegraphics[width=0.83\textwidth]{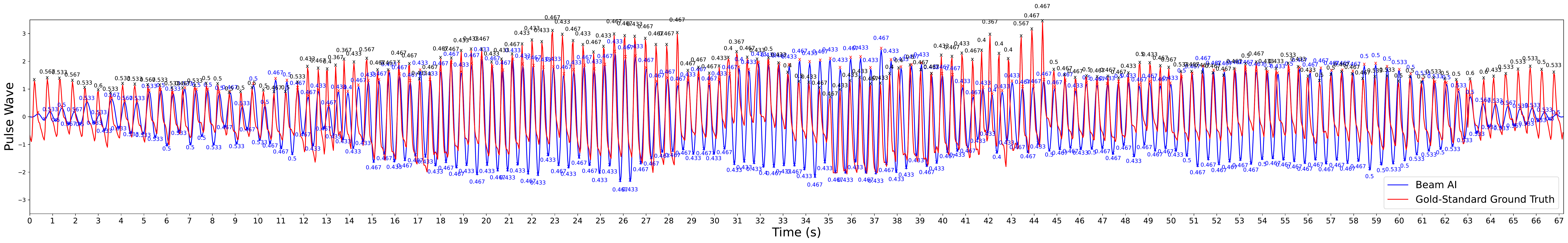} }}
    \caption{Five sample pulse waves produced by the Beam AI SDK (red) compared to the corresponding gold-standard pulse-wave simultaneously extracted by a medically approved pulse sensor (blue). All samples are from the UBFC \cite{Bobbia17_UBFC} benchmark of videos at 30fps.}
    \label{fig:ubfc-pulse-waves}
    \vspace{-0.25in}
\end{figure}

\subsubsubsection{\textbf{3.2.2 \hspace{0.075in} Evaluation on the MMSE-HR \cite{Zhang16_MMSE} Benchmark}}

\textbf{Benchmark:} The MMSE-HR \cite{Zhang16_MMSE} benchmark consists of a dataset of 102 videos from 40 subjects recorded at 1040x1392 raw resolution at 25fps. During the recordings, various stimuli such as videos, sounds and smells are introduced to induce different emotional states in the subjects. The ground truth waveform for MMSE-HR is the blood pressure signal sampled at 1000Hz. The dataset contains a diverse distribution of skin colors in the Fitzpatrick scale (II=8, III=11, IV=17, V+VI=4).

\textbf{Ground Truth Signal:} Following \cite{EfficientPhys_Liu2021}, the blood pressure signal for each video is bandpass filtered with a 2nd-order Butterworth \cite{Selesnick96_Butterworth} filter to exclude frequencies outside of 0.75Hz (corresponding to 45 beats per minute) and 2.5Hz (corresponding to 150 beats per minute). Then, the dominant frequency in the signal is extracted using Fast Fourier Transform (FFT) \cite{Brigham67_FFT} and subsequently multiplied by 60 to produce a ground truth heart rate reading in beats per minute. Note that the dominant frequency in a pulse signal is indeed the heart rate and is extracted by this procedure.

\textbf{Predictions by the Beam AI SDK:} For each video, the Python implementation of the ``Pulse Processor'' from our SDK estimates a high-quality pulse wave signal. However, for fairness in comparison to baselines, this raw pulse wave signal is not processed through the subsequent modules. Instead, we employ the standard FFT-based procedure to extract the ground truth readings (also used in competing methods). First, the signal is bandpass filtered with a 2nd-order Butterworth \cite{Selesnick96_Butterworth} filter to exclude frequencies outside of the 0.75Hz (corresponding to 45 beats per minute) to 2.5Hz (corresponding to 150 beats per minute) range. Then, the dominant frequency in each signal is extracted using Fast Fourier Transform (FFT) \cite{Brigham67_FFT} and multiplied by 60 to produce the heart rate reading in beats per minute for the video.

\textbf{Results:} As shown in Table \ref{tab:liu-eval-mmse}, the Beam AI SDK achieves a nearly 2x improvement over competing methods. Specifically, it establishes an MAE of 1.72 beats per minute, an MAPE of 2.24\%, and an RMSE of 4.03, indicating an average accuracy within 2 beats per minute of the gold-standard ground truth readings. Additionally, we observe a Pearson correlation of 0.95, indicating strong matching between increasing and decreasing trends in the ground truth and the predicted heart rate values.

\subsection{Evaluation on the UBFC Benchmark with Manually Hand Verified Pulse Peaks}
\label{sec:ubfc-clean}

\begin{table}[t]
    \centering
    \tabcolsep=0.5cm
    \begin{tabular}{lrrrr}
        {} & \multicolumn{4}{c}{\textbf{Beam AI Internal Data (30 fps)}} \\
        Reading & MAE$\downarrow$ & MAPE$\downarrow$ & RMSE$\downarrow$ & $\rho$ $\uparrow$ \\
        \midrule
        Heart Rate {\small (Beats Per Minute)} & 1.046 & 1.48\% & 1.471 & 0.959 \\
        Heart Rate Variability {\small (SDNN ms)} & 12.003 & 13.52\% & 16.444 & 0.781 \\ 
        Stress {\small (Baevsky Stress Index)} & 0.171 & 31.61\% & 0.243 & 0.850 \\
    \end{tabular}
    \vspace{0.05in}
    \caption{Evaluating stress, heart rate, and heart rate variability estimation from the subject's face on Beam AI's internal data. Ground truth readings are provided by a Polar H10 chest.}
    \label{tab:beam-ai-internal-benchmark}
    \vspace{-0.2in}
\end{table}

\textbf{Benchmark:} We continue to use the UBFC \cite{Bobbia17_UBFC} video dataset for this section. Please refer to Section \ref{sec:ubfc} for details on the video dataset and the synchronized pulse wave signal.

\textbf{Ground Truth Signal:} When comparing to prior literature in Section \ref{sec:ubfc}, we follow the procedure of Liu et al. \cite{EfficientPhys_Liu2021} for generating ground truth heart rate readings to assure consistency in comparison. However, after direct examination, it's clear that there are limited but non-zero instances where the standard peak detector of Scipy \cite{2020SciPy-NMeth} generates false pulse peaks. This prompted us to manually examine the pulse wave signal for every video in the UBFC benchmark and identify the peaks by hand \footnote{This was completed by Peyman Bateni without official direct medical training to perform peak detection.}. This ensures that they are accurately localized and can be used for stress and HRV evaluation. We then employ these hand-verified peaks to extract inter-beat intervals that are then used to calculate ground truth heart rate (beats per minute), heart rate variability (SDNN ms), and stress (according to the Baevsky Stress Index).

\textbf{Predictions by the Beam AI SDK:} For each video in the benchmark, the Python implementation of the ``Pulse Processor'' in our SDK estimates a high-quality pulse wave signal. We then generate the inter-beat intervals using an equivalent Python implementation of the  ``Inter-Beat Interval Processor'' from the SDK. The output inter-beat intervals are then directly used to estimate heart rate, heart rate variability, and stress for each video using the metric definitions inside the ``Biometric Estimator''.

\textbf{Results:} As demonstrated in Table \ref{tab:ubfc-eval-with-manual-data}, we achieve an MAE of 0.318 beats per minute (MAPE of 0.32\%) on heart rate estimation, achieving near-perfect pulse estimation on the majority of videos in the benchmark. Furthermore, we achieve an MAE of 11.125 ms (MAPE 20.26\%) with a high correlation score of 0.841, demonstrating the ability to produce commercially useful heart rate variability estimates that strongly correlate with the increases and decreases in the ground truth heart rate variability. Lastly, we achieve an MAE of 0.973 (MAPE of 44.43\%) on stress estimation. Despite a comparatively larger error rate, we demonstrate a strong correlation with ground truth readings, achieving a Pearson correlation score of 0.730.

\subsection{Evaluation on Beam AI's Internal Data}
\label{sec:iphone-data}

\begin{figure}[t]
    \centering
    \subfloat[heart rate Estimation over 60s Windows]{{\includegraphics[width=0.88\textwidth]{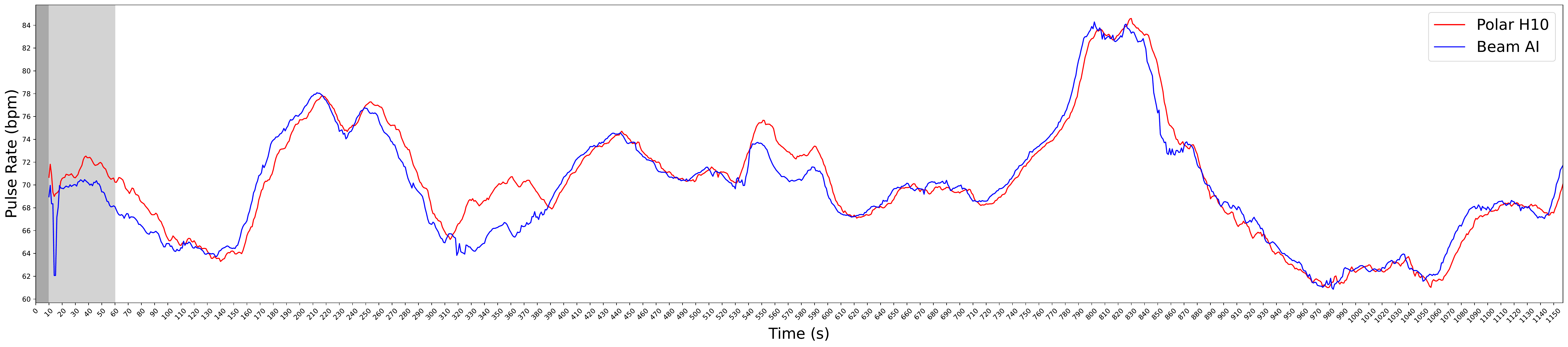} }} \\
    \vspace{-0.1in}
    \subfloat[HRV Estimation over 60s Windows]{{\includegraphics[width=0.88\textwidth]{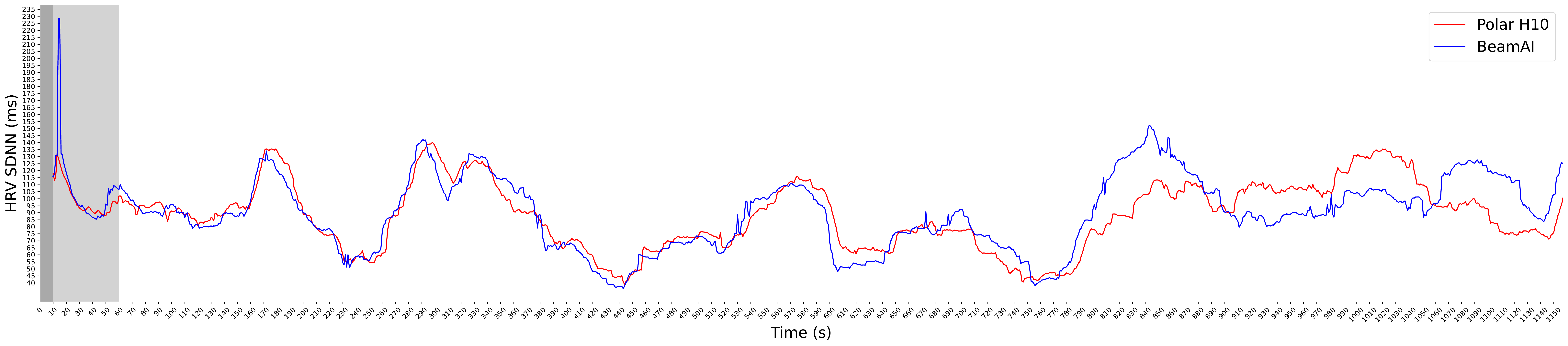} }} \\
    \vspace{-0.1in}
    \subfloat[Stress Estimation over 60s Windows]{{\includegraphics[width=0.88\textwidth]{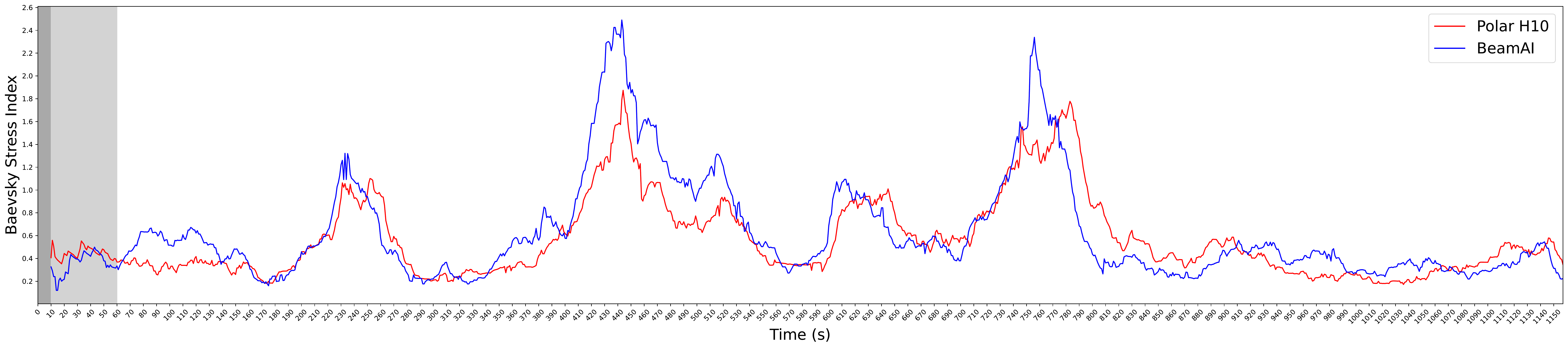} }}
    \\
    \vspace{-0.1in}
    \subfloat[Heart Rate Estimation Compared to Apple Watch over 20s Windows]{{\includegraphics[width=0.9\textwidth]{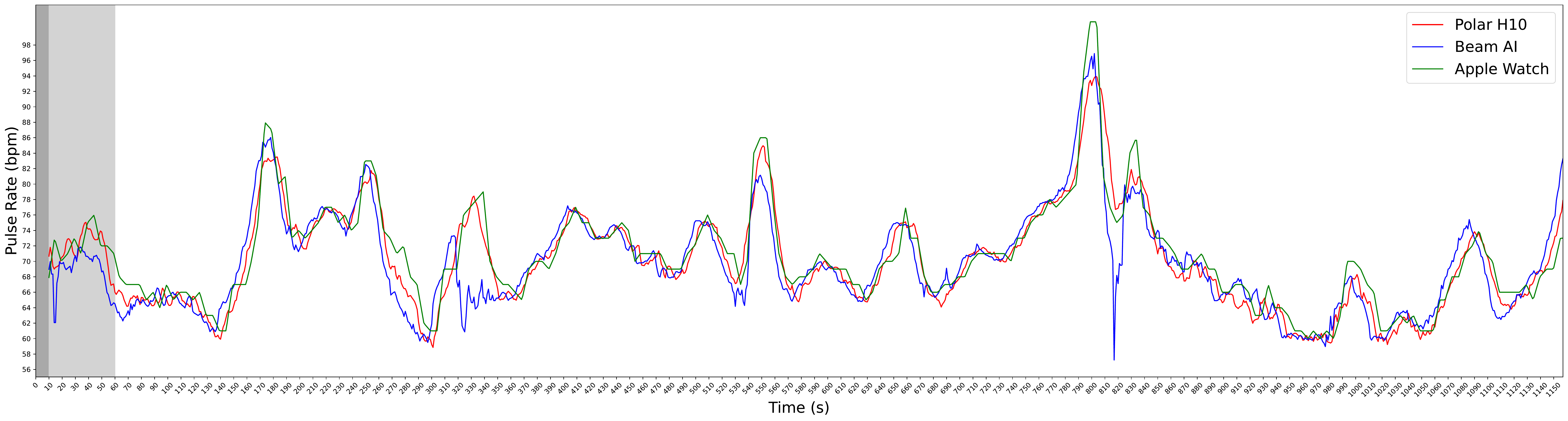} }}
    \caption{Continuous stress, heart rate and heart rate variability monitoring on Beam AI's internal data consisting of a 20-minute passive recording of Peyman Bateni on an iPhone 13 device. A sliding 60-second time window is used to estimate each value. Note that there is a 10-second minimum window for estimating values (shown in dark grey) and values before the 60s mark do not have a complete 60-second window (shown in light grey). In (d), we compare heart rate estimation over shorter 20-second windows to provide a comparison to measurements from a series 7 Apple Watch.}
    \label{fig:beam-ai-data}
    \vspace{-0.2in}
\end{figure}

\textbf{Benchmark:} We further evaluate the Beam AI SDK directly on Beam AI's internal data consisting of a 20-minute recording of a single subject (Peyman Bateni) on an iPhone 13 device. In the first half of the recording, the subject was holding the iPhone 13 device in a natural pose and using the device. In the second half, the device was placed next to the computer at the subject's workstation where he then proceeded to work on the computer for the remainder of the recording. The entire recording encompasses 19 minutes and 20 seconds and covers a range of movements when sitting and changes in lighting due to the large monitor in front of the subject.

\textbf{Ground Truth Signal:} A Polar H10 chest strap is worn by the subject during the recording from which, the subject's pulse in the form of inter-beat intervals is recorded using the EliteHRV iOS app \cite{elite_hrv_2022}. The inter-beat interval data is synchronized according to the recording's per frame time stamps and used to estimate stress, heart rate and heart rate variability over a moving 60s window according to their respective formulas from Section \ref{sec:technology}. Note that the 60s window is not fully complete during the first minute of the recording where naturally a 60s window is not available. As a result, there are no estimates for the first 10s and, in the 50s thereafter, we use the largest window size available (i.e. 10s window at 10th second, 30s window as 30th second, etc.).

\textbf{Predictions by the Beam AI SDK:} A pulse wave signal was extracted and recorded by the Beam AI SDK during the recording. This signal was saved and then post-processed using a Python implementation of our ``Inter-Beat Interval Processor'' module, resulting in a set of inter-beat intervals that are similarly grouped over 60s windows, with the exception of the first minute, where the same adaptive window strategy is used for measurements between the 10s and the 60s mark. The inter-beat intervals over each window are then subsequently used to produce continuous readings for the subject's heart rate, heart rate variability, and stress.

\textbf{Results:} We report overall performance on Beam AI's internal data in Table \ref{tab:beam-ai-internal-benchmark} and also provide graphs of continuous estimates over time for heart rate, heart rate variability and stress in Figure \ref{fig:beam-ai-data}. As shown, our technology achieves strong results, with an MAE of 1.046 beats per minute, 12.003 ms, and 0.171 for heart rate, heart rate variability and stress respectively. Furthermore, as indicated by the Pearson correlations achieved and shown in Figure \ref{fig:beam-ai-data}, we are able to estimate values that strongly correlate with the gold-standard ground truth from the Polar H10 chest-strap monitoring device.

\textbf{Comparison to Apple Watch:} We further evaluate our technology accuracy as compared to a series 7 Apple Watch that was worn simultaneously during the recording and used to extract heart rate values at every 5s intervals. These values were then interpolated to produce per-second heart rate estimates. Unfortunately, inter-beat intervals and continuously updated heart rate variability estimates are not available for third-party usage on the Apple Watch, and accordingly we cannot compare those. The results are shown in Figure \ref{fig:beam-ai-data}-d. Here, we reduce the window size for our estimates from 60s to the 20s window that is believed to be used by the Apple Watch. As shown, we are able to produce heart rate estimates that strongly correlate with the Apple Watch measurements. Overall, we achieve an MAE of 1.959 beats per minute whereas the Apple Watch achieves an MAE of 1.399 beats per minute when compared to the Polar H10 device. This indicates that we are approximately 0.6 beats per minute less accurate than the Apple Watch on average during seated phone usage.

\begin{table}[t]
    \centering
    \tabcolsep=1.67cm
    \begin{tabular}{lc}
        Model & Time To Process 1 Frame \\
        \midrule
        Beam AI & \textbf{0.5 ms} \\
        \midrule
        EfficientPhys-T1 \cite{EfficientPhys_Liu2021} & 30.0 ms \\
        TS-CAN \cite{TSCAN_Liu2020} & 6.0 ms \\
        EfficientPhys-C \cite{EfficientPhys_Liu2021} & 4.0 ms \\
        POS \cite{POS_Wang2016} & 2.7 ms \\
        CHROM \cite{CHROM_DeHaan2013} & 2.8 ms \\
        ICA \cite{ICA_Poh2011} & 3.1 ms \\
    \end{tabular}
    \vspace{0.05in}
    \caption{Processing speed on devices. Note that the Beam AI SDK was evaluated on an iPhone 13 device while baselines are reference run-times on an ARM CPU \cite{EfficientPhys_Liu2021, TSCAN_Liu2020}.}
    \label{tab:processing-speed}
    \vspace{-0.2in}
\end{table}

\subsection{Processing Speed}
\label{sec:processing-speed}

Inference on mobile devices is best done on the device as it preserves user privacy, can operate in real-time and reduces the rate of frame loss. However, this requires very efficient models to be able to run real-time processing, especially at high framerates. Table \ref{tab:processing-speed} compares the processing speed of the Beam AI SDK with competing methods. As shown, the Beam AI SDK takes 0.5 ms to process one frame, a near 6x improvement over the fastest competing methods. This enables the Beam AI SDK to run smoothly at 120fps while using limited computational resources on the device.

\section{Discussion}
\label{sec:discussion}

We introduce the Beam AI SDK to enable smartphone apps to monitor user stress in real-time. We provide two sample apps (Beam AI Lite and Beam AI Browser) on App Store to demonstrate some applications of real-time stress monitoring inside apps. We further establish the empirical efficacy of the Beam AI SDK by validating the underlying technology on UBFC \cite{Bobbia17_UBFC}, MMSE-HR \cite{Zhang16_MMSE} and Beam AI's internal data. We demonstrate nearly twice better accuracy as compared to competing methods while running up to six times faster on mobile devices.

\textbf{Acknowledgements:} We have conducted experiments with publicly available datasets and privately collected data. Our experiments are consistent with best published practices in the academic domain \cite{TSCAN_Liu2020, EfficientPhys_Liu2021, MetaPhys_Liu2021, ICA_Poh2011, POS_Wang2016}. However, we have not conducted medical grade testing with strict medical studies and guidelines to validate our measurements. For this reason, we cannot make any claims on medical reliability of our measurements or their relevance for any sort of medically-relevant diagnostics. This is something we will explore for future iterations of our application and deployments. In addition, this is why we have clear messaging whenever a recording is in progress in our demo apps that ``Beam AI is not medically approved and should not be used for health decisions'' as shown in Figure \ref{fig:screenshots}. We strongly recommend interested developers maintain the necessary disclaimer messaging when using the Beam AI SDK for applications that are intended for or can be mistaken for medical usage.

\textbf{Future Studies:} We are undertaking an extensive empirical study with a large set of participants in Vancouver. This will extensively evaluate our technology during diverse phone usage (such as video replay, gaming, texting, emailing, browsing, and social networking) in different lighting and motion settings. We will report these results publicly once the study completes.

\textbf{Future Directions:} The Beam AI SDK is currently available on iOS only. We will expand support to other mobile operating systems, cross-platform development frameworks, and desktop operating systems in the future. We are also developing improved core technologies for a more robust extraction of the user pulse wave in noisy environments.

\bibliographystyle{plainnat}
\bibliography{main}

\end{document}